# A Novel Grouping-Based Hybrid Color Correction Algorithm for Color Point Clouds

**Kuo-Liang Chung[1]**  · **Ting-Chung Tang[1]**

**Abstract** Color consistency correction for color point clouds is a fundamental yet important task in 3D rendering and compression applications. In the past, most previous color correction methods aimed at correcting color for color images. The purpose of this paper is to propose a grouping-based hybrid color correction algorithm for color point clouds. Our algorithm begins by estimating the overlapping rate between the aligned source and target point clouds, and then adaptively partitions the target points into two groups, namely the close proximity group $G^{cl}$ and the moderate proximity group $G^{mod}$, or three groups, namely $G^{cl}$, $G^{mod}$, and the distant proximity group $G^{dist}$, when the estimated overlapping rate is low or high, respectively. To correct color for target points in $G^{cl}$, a K-nearest neighbors based bilateral interpolation (KBI) method is proposed. To correct color for target points in $G^{mod}$, a joint KBI and the histogram equalization (JKHE) method is proposed. For target points in $G^{dist}$, a histogram equalization (HE) method is proposed for color correction. Finally, we discuss the grouping-effect free property and the ablation study in our algorithm. The desired color consistency correction benefit of our algorithm has been justified through 1086 testing color point cloud pairs against the state-of-the-art methods. The C++ source code of our algorithm can be accessed from the website: `https://github.com/ivpml84079/Point-cloud-color-correction`.

## 1 Introduction

Colored point clouds, in particular, are often considered as a fusion of depth information and colored images. With the flourishing development of 3D imaging sensors [13], there has been a progression from traditional methods, which generate stereoscopic images using image disparity maps and employ structure from motion to derive relative depth through disparity estimation [26], to more advanced techniques, such as the integration of structured light and color images in RGB-D cameras [42], as well as the direct acquisition of depth information through light field [32] and LIDAR cameras. Some key applications of colored point clouds include 3D vision, autonomous driving, remote sensing, object tracking, augmented virtual reality, archaeology, cultural heritage, and building modeling.

To encompass the entire scene covered by multi-view point clouds, different methods have been employed to solve matrix transformations. These transformations involve scaling, rotation, and translation operations to map the source point cloud onto the target domain, ensuring effective alignment between the source and target point clouds. In the past years, the iterative closest point (ICP) related registration methods [3], [4], [6], [19] and the random sample consensus (RANSAC) related registration methods [12], [36], [37], [17] have been developed. Without the need of an initial correspondence set, ICP iteratively refines the alignment by minimizing the distance between corresponding points in the aligned source and target point clouds, leading to fast execution time performance. With the need of an initial correspondence set, RANSAC is robust in registration accuracy due to the iterative hypothesis-and-verification (HAV) process, but RANSAC is time consuming.

However, when dealing with color point clouds, it must consider both feature and color attributes together. Men et al. [20] proposed a 4D ICP registration method to accelerate the registration process by using hue data. Then, the transforma-

✉ Kuo-Liang Chung
E-mail: klchung01@gmail.com
Ting-Chung Tang
E-mail: m11115097@mail.ntust.edu.tw

1 Department of Computer Science and Information Engineering, National Taiwan University of Science and Technology, No. 43, Section 4, Keelung Road, Taipei, 10672, Taiwan, R.O.C.

tion matrix solution is iteratively solved using the singular value decomposition (SVD) method. Park et al. [21] determined the transformation matrix by solving a joint photometric and geometric objective function that locks the alignment along both the normal direction and the tangent plane. Based on the supervoxel segmentation, Yang et al. [38] proposed a hybrid feature representation with color moments to build more robust relationships for solving the transformation matrix. Wan et al. [30] deployed the correntropy and bi-directional distance into color point cloud registration, where the correntropy is used to remove outliers and the bi-directional distance is used to avoid being trapped into local optimum. Wan et al. [31] proposed an effective registration method which combines three strategies: the salient object detection-based strategy, the bidirectional color distance guided strategy, and the outlier handling strategy. Considering the influence of background, Yao et al. [39] proposed a joint objective to align both salient color points and background points, improving the registration alignment accuracy.

## 1.1 The Importance of Color Correction for Color Point Clouds

Unfortunately, issues related to variations in luminance and chrominance often emerge practically after aligning two color point clouds. These inevitable problems can be attributed to differences in time and device parameters during capture, such as white balance and exposure time, as well as variations in lighting conditions resulting from differences in shooting angles. These varying factors collectively contribute to a perceptually unpleasant appearance in the alignment between the aligned source point cloud and the target point cloud. Furthermore, this color inconsistency problem can negatively impact various important applications, such as the color point cloud compression using chroma attributes [28], [27], color point cloud-based human pose estimation [44], [40], and point cloud-based rendering [16], [15].

To ensure the generation of composite point clouds that appear visually pleasing and coherent, color consistency correction, color correction for short, is important yet challenging for color point clouds. The purpose of this paper is to propose a novel color correction algorithm to harmonize color attributes across the aligned color point clouds, enhancing the overall visually coherent and color-consistent effects of the final composite.

## 1.2 Related Works

After examining the related color correction works published in the past, we found that most of the related color correction works focus on color images instead of the trending topic: color point clouds. Below, we provide an overview of the related color correction works, along with some of their advantages and disadvantages.

The nearest neighbors based (NN) method is straightforward and computationally efficient. The color of the target point is corrected by that of the nearest neighbor in the aligned source point cloud. The NN method may not handle color variations well, leading to potential color mismatches in regions with sparse point density. The K-nearest neighbors based (KNN) method [1] is a simple but more effective method relative to the NN method. For each target point $p^t$, the K-nearest aligned source points of $p^t$ are first searched. Next, the average color of the found K-nearest aligned source points of $p^t$ is used to replace the color of the color of $p^t$. Although the KNN method is simple, it suffers from an unpleasant perception artifact when the distance between $p^t$ and $p^{s,ali}$ increases.

In the histogram matching based (HM) method [11], the cumulative distribution functions [24] of the target point cloud and the aligned source point cloud are first built up. Let the two cumulative distribution functions be denoted by $F^t$ and $F^{s,ali}$, respectively. Based on $F^t$ and $F^{s,ali}$, a histogram mapping function is constructed. Taking the color of each target point $p^t$ as a key, the color of $p^t$ is replaced by the mapped color of the histogram mapping function. It can effectively handle global color variations and ensure consistent color distributions. However, it may not capture fine-grained color details.

In the auto-adapting global-to-local based (AGL) method [41], it begins by incorporating the preservation of mean and standard deviation values into a linear color mapping model. Next, using a least squares technique to solve the color mapping model, it can ensure that the aligned source point cloud and the color-corrected target point cloud have globally similar mean and standard deviation values for their color distributions. However, it may not preserve fine-grained color details.

To improve the HM method, the hybrid HM-based (HHM) method [8] was proposed. The histograms of both the target and aligned source point clouds are sorted separately based on the frequencies of their bins, and then it arranges the bins in increasing order of frequency. Further, the two corresponding cumulative distribution functions are used to create a tentative histogram mapping function, and then it continuously removes nearly vertical and horizontal segments in the mapping function until the accumulated frequency of the removed segments reaches 5%. The removed segments are replaced by a uniform distribution.

The related color correction works introduced above are used for color images, particularly on RGB-color consistency correction. Recently, based on the generated 6-D point clouds which are estimated using the depth information captured by a time-of-flight sensor and the RGB sensor, the point cloud color constancy (PCCC) method [35] adopted the modified variant of PointNet [22] to estimate the illumination chromaticity. In PCCC, it consists of one block for extracting point-wise features and the other block for estimating the spatial weighted illuminant. It is noticeable that different from the above-mentioned related works focusing on RGB-color consistency correction, the PCCC method focuses on illumination chromaticity consistency correction.

### 1.3 The outline of our research methods and contributions

Given a source point cloud $P^s$ and a target point cloud $P^t$, taking the overlapping rate of the point cloud pair $(P^s, P^t)$ and the classification of target points into account, in this paper, we propose a novel grouping-based hybrid color correction algorithm for color point clouds. As depicted in Fig. 1, the outline of our research methods and contributions are described below.

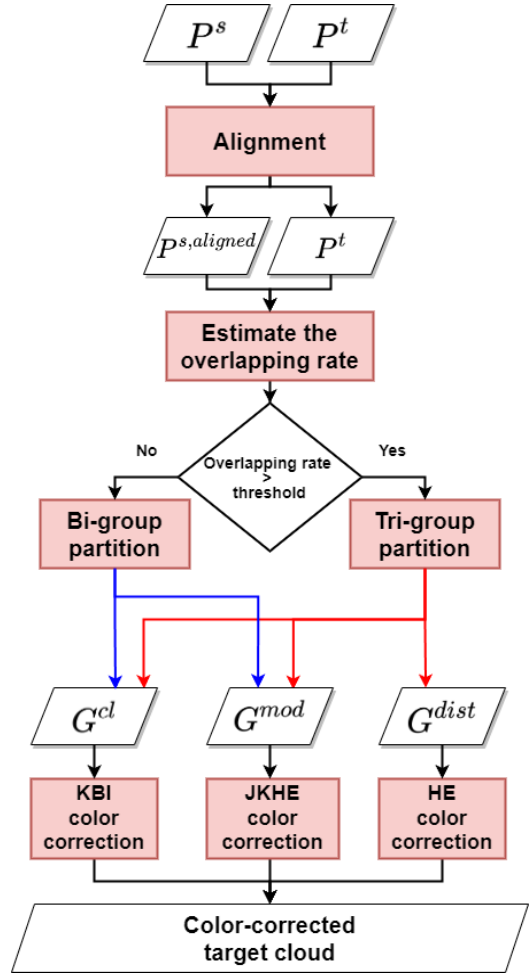

**Fig. 1** The pipelines of our color correction algorithm for color point clouds.

- Different from the PCCC method [35] and the related color correction works for color images, after registering $P^s$ and $P^t$, the overlapping rate between the aligned source point cloud and the target point cloud is estimated by taking the distance between each target point and the nearest aligned source point into account. If the estimated overlapping rate is less than the specified threshold $T_r$, it indicates that there is a certain ratio of pairs between the aligned source points and the nearest target points belong to the close proximity group $G^{cl}$, while the rest pairs belong to the moderate group $G^{mod}$. Therefore, all target points are partitioned into $G^{cl}$ and $G^{mod}$. Otherwise, if the estimated overlapping rate is greater than or equals $T_r$, all target points are partitioned into three groups, namely $G^{cl}$, $G^{mod}$, and the distant proximity group $G^{dist}$.

- Because the distance between each target point $p^t$ in the close proximity group $G^{cl}$ and its nearest aligned source point is less than that in the other groups, $G^{mod}$ and $G^{dist}$, the color of $p^t (\in G^{cl})$ is corrected by using the proposed K-nearest neighbors based bilateral interpolation (KBI) method. In the proposed KBI method, only valid local reference aligned source points are used to correct the color of $p^t$ for achieving a better color correction effect. Furthermore, to correct color for each target pixel $p^t \in G^{mod}$, the joint KBI and the histogram equalization (JKHE) method is proposed because the distance between $p^t$ and its nearest aligned source point is relatively moderate relative to the other two groups. To correct color for each target point $p^t$ in $G^{dist}$, a global histogram equalization(HE) method is proposed because the distance between $p^t$ and its nearest aligned source point is relatively distant relative to the other two groups. In addition, the grouping-effect free property is provided.

–  In the large-scale testing color point cloud pairs, 146 point cloud pairs are adopted from the SHOT dataset [25] and 940 point cloud pairs are adopted from the augmented ICL-NUIM dataset [7]. Comprehensive experiments have been carried out to justify the quantitative and qualitative quality merits of our algorithm when compared with the state-of-the-art methods: the NN method, the KNN method [1], the HM method [11], the AGL method [41], and the HHM method [8].

The remainder of this paper is organized as follows. In Section 2, the proposed method is presented to partition the target points into two groups or three groups. In Section 3, the proposed grouping-based hybrid algorithm is presented to correct color for target point clouds. In Section 4, comprehensive experimental data are demonstrated to show the quantitative and qualitative quality merits of our algorithm. In Section 5, some concluding remarks are presented.

## 2 The proposed method for partitioning target points into two or three groups

### 2.1 The purpose of group partition

The purpose of the partition of all target points is that the reference color information to correct color for target points applied to each group depends on their proximity to the corresponding aligned source points. For example, each target point $p^t$ in $G^{cl}$ could utilize some local color difference information from the neighboring aligned source points to correct the color of $p^t$. While each target point $p^t$ in $G^{dist}$ could utilize far and global information between the target points in $P^t$ and the aligned source points in $P^{s,ali}$ to correct the color of $p^t$.

### 2.2 Our partition method

To better understand our partition method, one high overlapping alignment example and one low overlapping alignment example are taken to explain how our partition method works.

Considering a high overlapping alignment example, Fig. 2(a) and Fig. 2(b) show the source and target point clouds, respectively, and Fig. 2(c) shows the alignment. In Fig. 2(c), the Euclidean distance between each target point and its nearest aligned source point is calculated. Fig. 2(d) illustrates the distance distribution of the alignment, where the x-axis represents the distance parameter and the y-axis represents the frequency.

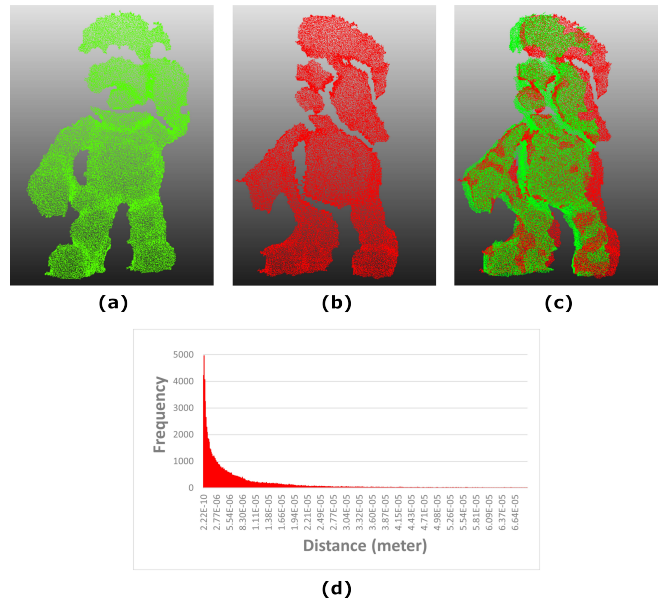

**Fig. 2**  The distance distribution between the aligned source point cloud and the target point cloud. (a) The source point cloud. (b) The target point cloud. (c) The alignment. (d) The distance distribution of the alignment.

### 2.2.1 Partitioning the target points into three groups for high overlapping alignment

Based on the distance distribution of the alignment in Fig. 2(d), we now present how to estimate the overlapping rate of the alignment in Fig. 2(c). If the distance between $p^t$ and $P^{s,ali}$ is less than the specified threshold $T_d$, it donates one vote to the point pair $(p^t, P^{s,ali})$. After accumulating the votes of all point pairs in the alignment, the overlapping rate of the alignment is estimated as $R = \frac{V}{|P^t|}$ where $V$ denotes the accumulated votes and $|P^t|$ denotes the cardinality of $P^t$. For the alignment in Fig. 2(c), the estimated overlapping rate is 67.401%.

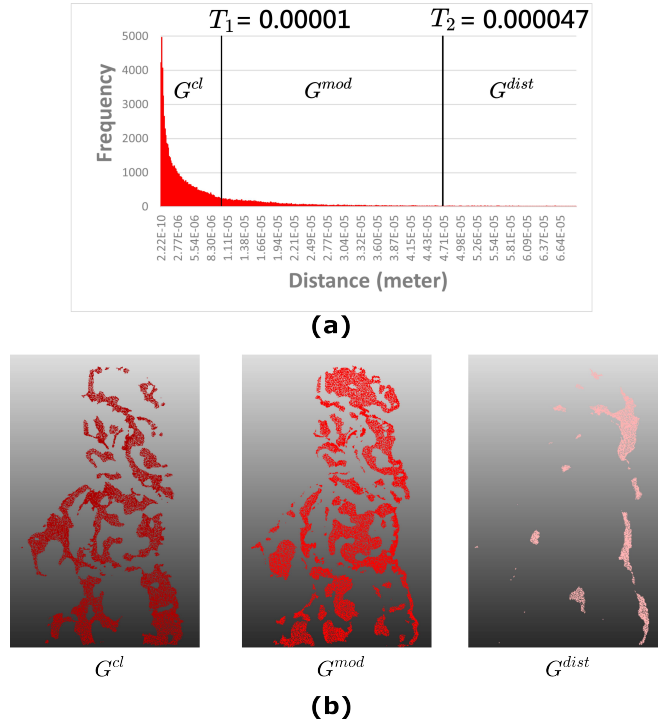

**Fig. 3** Tri-group partition for the high overlapping alignment example in Fig. 2(c). (a) The determined two thresholds, $T_1$ and $T_2$. (b) The partitioned three target groups.

Because the estimated overlapping rate "67.401%" is larger than the specified threshold $T_r$, which is set to 45%, Fig. 2(c) is a high overlapping alignment, we adopt the three-level thresholding method [18] to partition the distance distribution in Fig. 2(d) into three groups, where the determined two thresholds are $T_1$ (= 0.00001) and $T_2$ (= 0.000047) which are depicted in Fig. 3(a). Based on the two threshold values, $T1$ and $T2$, the target points in $P^t$ are partitioned into three groups: $G^{cl}$, $G^{mod}$, and $G^{dist}$, as depicted in Fig. 3(b).

### 2.2.2 Partitioning the target points into two groups for low overlapping alignment

Considering an alignment example in Fig. 4(a), where the overlapping rate is 44.49%, Fig. 4(b) shows the distance distribution of this alignment. Because the overlapping rate of Fig. 4(a) is less than the specified threshold $T_r$ (= 45%), the alignment in Fig. 4(a) is a low overlapping alignment. Therefore, following the proposed group partition strategy in Fig. 1, the target points in $P^t$ are partitioned into two groups: $G^{cl}$ and $G^{mod}$. To this end, we adopt the two-level thresholding method [5] to partition the distance distribution of Fig. 4(b) into two parts, where the determined threshold $T_b$ (= 0.0000904877) is depicted in Fig. 4(c). Based on the threshold value $T_b$, the target points in $P^t$ are partitioned into two groups: $G^{cl}$ and $G^{mod}$, as depicted in Fig. 4(d).

## 3 Our color correction algorithm

Based on the pipelines of our color correction algorithm, our algorithm, Algorithm 1 for short, is listed below.

---

**Algorithm 1:** Our Color Correction Algorithm

---

**Input:** The aligned source color point cloud $P^{s,ali}$ and the target point cloud $P^t$.

**Output:** The color-corrected target point cloud.

**Step 1:** Estimate the overlapping rate $R$ between $P^t$ and $P^{s,ali}$.

**Step 2:** If $R$ is less than or equal to the threshold $T_r$, go to Step 3; otherwise, apply the tri-group partition method to partition all target points into three groups. Correct color for each target point in $G^{cl}$, $G^{mod}$, and $G^{dist}$ by using the KBI, JKHE, and HE methods, respectively. Go to Step 4.

**Step 3:** Apply the bi-group partition method to partition all target points into two groups. Correct color for each target point in $G^{cl}$ and $G^{mod}$ by using the KBI and JKHE methods, respectively.

**Step 4:** Report the color-corrected target points.

---

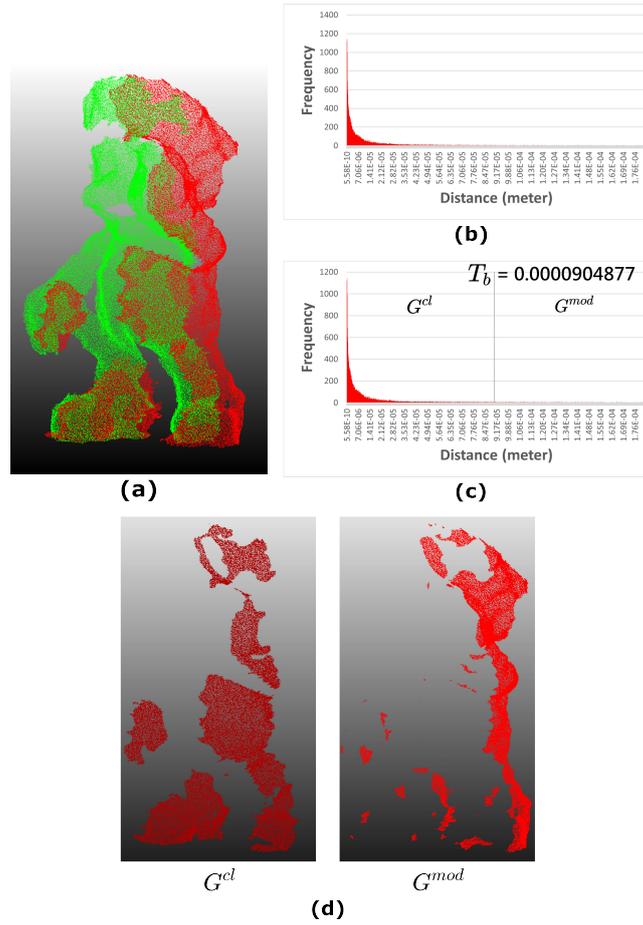

**Fig. 4** Bi-group partition for a low overlapping alignment example. (a) The alignment. (b) The distance distribution. (c) The determined threshold $T_b$. (d) The partitioned two target groups.

In what follows, the proposed KBI, JKHE, and HE methods used in Algorithm 1 for correcting color for target points in $G^{cl}$, $G^{mod}$, and $G^{dist}$, respectively, are presented in Subsection 3.1, Subsection 3.2, and Subsection 3.3 for details. In Subsection 3.4, the grouping-effect removal advantage of our color correction algorithm is discussed.

### 3.1 Correct color for target points in $G^{cl}$ via our KBI method

As mentioned before, because each target point $p^t$ in $G^{cl}$ is relatively close to its nearest aligned source point, we propose the KBI method to correct color for $p^t$ using a valid local reference aligned source point set $\mathcal{N}(p^t)$ which denotes the set of the K-nearest aligned source points of $p^t$.

#### 3.1.1 Determining the valid local reference aligned source points of each target point in $G^{cl}$

Although the target point $p^t$ in $G^{cl}$ is relatively close to its K-nearest aligned source points $\mathcal{N}(p^t)$, some invalid aligned source points in $\mathcal{N}(p^t)$, whose colors are much different from the target point $p^t$, may still be included as the reference points, degrading the color correction performance. To remove these invalid reference aligned source points in $\mathcal{N}(p^t)$, for each target point $p^t \in G^{cl}$, we adopt the CIELAB color difference constraint [23] to remove the invalid aligned source points in $\mathcal{N}(p^t)$. When the LAB color difference between the target point $p^t$ and the reference aligned source point $P^{s,ali} \in \mathcal{N}(p^t)$ is larger than the specified threshold, the invalid aligned source point $P^{s,ali}$ is deleted, resulting in the valid local reference aligned source point set $\mathcal{N}'(p^t)$.

As a result, the c-color, $c \in \{R, G, B\}$, difference between the target point $p^t$ and each point $p' \in \mathcal{N}'(p^t)$ is defined by

$$D^c(p', p^t) = C^c(p') - C^c(p^t) \qquad (1)$$

where $C^c(p')$ and $C^c(p^t)$ denote the c-color values of $p'$ and $p^t$, respectively.

### 3.1.2 Correct color of each target point in $G^{cl}$ via the KBI method

For each target point $p^t$ in $G^{cl}$, with the help of its valid reference aligned source points in $\mathcal{N}'(p^t)$ and the KBI color correction method, the c-color of $p^t$ is corrected by

$$
\begin{aligned}
C^c(p^t) &= C^c(p^t) + f^{c,KBI}(p^t, \mathcal{N}'(p^t)) \\
&= C^c(p^t) + \frac{1}{\sum_{p' \in \mathcal{N}'(p^t)} w_{p'}} * \sum_{p' \in \mathcal{N}'(p^t)} w_{p'} * D^c(p', p^t)
\end{aligned}
\tag{2}
$$

with

$$
w_{p'} = \exp(-\frac{||d(p^t, p')||^2}{\sigma_d^2}) * \exp(-\frac{||D^c(p', p^t)||^2}{\sigma_c^2})
\tag{3}
$$

where $D^c(p', p^t)$ has been defined in Eq. 1. $d(p^t, p')$ denotes the Euclidean distance between $p^t$ and the reference point $p' \in \mathcal{N}'(p^t)$. $\sigma_d$ and $\sigma_c$ are the specified distance and color difference deviations.

After correcting color for all target points in the close proximity group $G^{cl}$ using the proposed KBI method, we proceed to correct color for all target points in the moderate proximity group $G^{mod}$.

## 3.2 Correct color for target points in $G^{mod}$ via our JKHE method

For each target point $p^t$ in the group $G^{mod}$, we propose a joint KBI and HE (JKHE) method to correct $p^t$, in which the KBI method refers to some valid local aligned source points, while the HE method refers to all target points in $G^{cl}$ and $G^{mod}$, denoted by the target point set $P^{t,HE}$, and the corresponding reference aligned source point set of $P^{t,HE}$, denoted by $P^{s,ali,HE}$. Finally, an effective fusion strategy is proposed to fuse the color correction results from the KBI method and the HE method for correcting the color of $p^t$.

### 3.2.1 Correct color of each target point in $G^{mod}$ via the JKHE method

Using the proposed JKHE method, the color of $p^t$ is corrected by

$$
\begin{aligned}
C^c(p^t) = C^c(p^t) &+ w_1 * f^{c,KBI}(p^t, \mathcal{N}'(p^t)) \\
&+ w_2 * f^{c,HE}(p^t)
\end{aligned}
\tag{4}
$$

where $1 = w_1 + w_2$. The weight $w_1$ denotes the weight assigning to the KBI color correction function $f^{c,KBI}(p^t, \mathcal{N}'(p^t))$ (see Eq. 2). The weight $w_2$ denotes the weight assigning to the HE color correction function $f^{c,HE}(p^t)$, which will be presented in Subsection 3.2.2 in detail. In Subsection 3.2.3, the determination of the two weights, $w_1$ and $w_2$, will be presented in detail.

### 3.2.2 The derivation of the HE color correction function $f^{c,HE}(p^t)$ used in Eq. 4

To derive the HE color correction function $f^{c,HE}(p^t)$ in Eq. 4 for realizing the HE method, the working target point set $P^{t,HE}$ is selected as $G^{cl} \cup G^{mod}$; the corresponding reference aligned source point set of $P^{t,HE}$ is denoted by $P^{s,ali,HE}$. Based on $P^{t,HE}$ and $P^{s,ali,HE}$, we construct two c-color histograms, $H^{c,t,HE} = \{h^{c,t,HE}(i_k)|0 \le k \le 255\}$ and $H^{c,s,ali,HE} = \{h^{c,s,ali,HE}(i_k)|0 \le k \le 255\}$ for $c \in \{R, G, B\}$, respectively, where the x-axis denotes the c-color value over $[0, 255]$ and the y-axis denotes the frequency. Here, the bin $h^{c,t,HE}(i_k)$ represents the number of the target points with c-color value $i_k$ and the bin $h^{c,s,ali,HE}(i_k)$ represents the number of the aligned source points with c-color value $i_k$.

Based on the histogram equalization technique [14], we first utilize $H^{c,t,HE}$ and $H^{c,s,ali,HE}$ to construct two cumulative c-color histograms, $H'^{c,t,HE}$ and $H'^{c,s,ali,HE}$, respectively, where

$$
H'^{c,t,HE}(i_k) = \sum_{j=0}^{k} h^{c,t,HE}(i_j)
$$

$$
H'^{c,s,ali,HE}(i_k) = \sum_{j=0}^{k} h^{c,s,ali,HE}(i_j)
\tag{5}
$$

Using the two cumulative c-color histograms, $H'^{c,t,HE}$ and $H'^{c,s,ali,HE}$, the corrected c-color value of $p^t$ in $G^{mod}$ is expressed by

$$u = C^c(p^t) + f^{c,HE}(p^t)$$
$$\text{with}$$
$$H^{\prime c,s,ali,HE}(u) \leq H^{\prime c,t,HE}(C^c(p^t)) < H^{\prime c,s,ali,HE}(u+1)$$

(6)

where the color correction function $f^{c,HE}(p^t)$ equals $u - C^c(p^t)$, where $C^c(p^t)$ denotes the original c-color value of the target pixel $p^t$.

### 3.2.3 Determining the two weights: $w_1$ and $w_2$

In the group $G^{mod}$, for each target point $p^t$, we first calculate the average distance between the target point $p^t$ and its valid reference aligned source points in $\mathcal{N}'(p^t)$, and the calculated average distance is denoted by $d^{avg}(p^t)$. Next, all of these $|G^{mod}|$ average distances are sorted in increasing order. Let the target points corresponding to the sorted average distance sequence be denoted by the ordered set $\langle p^t_{1'}, p^t_{2'}, ..., p^t_{|G^{mod}|'} \rangle$, where in the ordered target point set, the average distance between the first target point $p^t_{1'}$ and its valid reference aligned source points in $\mathcal{N}'(p^t_{1'})$, denoted by $d^{avg}(p^t_{1'})$, is minimal among the $|G^{mod}|$ average distances.

In the sequence $\langle p^t_{1'}, p^t_{2'}, ..., p^t_{|G^{mod}|'} \rangle$, because the average distance of the target $p^t_{|G^{mod}|'}$ is maximal, the HE color correction function $f^{c,HE}$ dominates the color correction of $p^t_{|G^{mod}|'}$ but the KBI color correction function $f^{c,KBI}$ is insignificant to correct color for $p^t_{|G^{mod}|'}$. On the contrary, because the average distance of the target $p^t_{1'}$ is minimal, the KBI color correction function dominates the color correction of $p^t_{1'}$ but the HE color correction function is insignificant to correct color for $p^t_{1'}$. The following observation is provided.

**Observation 1:** In the moderate group $G^{mod}$, for each target point $p^t$, we observe that the smaller the average distance between $p^t$ and its valid local reference aligned source points is, the more important the KBI color correction function $f^{c,KBI}(p^t, \mathcal{N}'(p^t))$ is; that is, the larger the weight $w_1$ assigned to $f^{c,KBI}(p^t, \mathcal{N}'(p^t))$ is.

According to Observation 1, the weights $w_2$ and $w_1$ are defined by

$$w_2 = \frac{d^{avg}(p^t) - d^{avg}(p^t_{1'})}{d^{avg}(p^t_{|G^{mod}|'}) - d^{avg}(p^t_{1'})}$$
$$w_1 = 1 - w_2$$

(7)

After correcting color for all target points in the moderate proximity group $G^{mod}$ by Eq. 4 for the bi-group case or the tri-group case (see Fig. 1), in the next subsection, we describe how to correct color for target points in the distant proximity group $G^{dist}$.

### 3.3 Correct color for target points in $G^{dist}$ via the HE method

Because the distance between each target point $p^t$ in the distant proximity group $G^{dist}$ and its nearest aligned source point is larger than that in the two groups, $G^{cl}$ and $G^{mod}$, we propose the HE method to correct color for $p^t$ in $G^{dist}$. To derive the color correction function $f^{c,distant,HE}(p^t)$ for each target point in $G^{dist}$, the working target point set $P^{t,HE}$ is selected as $G^{cl} \cup G^{mod} \cup G^{dist}$; the corresponding reference aligned source point set of $P^{t,HE}$ is denoted by $P^{s,ali,HE}$.

Similar to the c-color histogram construction for the group $G^{mod}$, we construct the two c-color histograms, $H^{c,dist,t,HE}$ and $H^{c,dist,s,aligned,HE}$, using the working target point set $P^{t,HE}$ and the corresponding reference aligned source point set $P^{s,ali,HE}$. In addition, based on the two constructed c-color histograms, we further construct two cumulative c-color histograms, $H^{\prime c,dist,t,HE}$ and $H^{\prime c,dist,s,ali,HE}$. Using $H^{\prime c,dist,t,HE}$ and $H^{\prime c,dist,s,ali,HE}$, the corrected c-color value of the target point $p^t$ is expressed by

$$u = C^c(p^t) + f^{c,HE}(p^t)$$
$$\text{with}$$
$$H^{\prime c,dist,s,ali,HE}(u) \leq H^{\prime c,dist,t,HE}(C^c(p^t)) < H^{\prime c,dist,s,ali,HE}(u+1)$$

(8)

### 3.4 The grouping-effect free property discussion

In this subsection, the grouping-effect free property in the proposed algorithm is discussed. The grouping-effect free property leads to the removal of grouping artifacts at the boundary of two adjacent target groups. The grouping-effect free property at the boundary of two adjacent target groups is defined below.

**Definition 1**  At the boundary of two adjacent target groups, if the color of the leftmost target point in the right target group is corrected by the same color correction method for the rightmost target point in the left target group, it is said that the color-corrected target points at the boundary are grouping-effect free.

According to Definition 1, the grouping-effect free property of our algorithm is shown below.

**Proposition 1**  *Our color correction algorithm has the grouping-effect free property.*

*Proof*  We first consider the target points at the boundary between $G^{cl}$ and $G^{mod}$. Because the color of the rightmost target point in $G^{cl}$ is corrected by the KBI method, the proving key is to show whether the color of the leftmost target point in $G^{mod}$ is corrected by the KBI method too.

As mentioned in subsection 3.2.3, in the sorted target point sequence $\langle p_{1'}^t, p_{2'}^t, ..., p_{|G^{mod}|'}^t \rangle$, the average distance between the leftmost target point $p_{1'}^t$ and its valid reference aligned source points in $\mathcal{N}'(p^t)$ is minimal among the $|G^{mod}|$ average distances. From Eq. 7, it leads to $w_2 = 0$ and $w_1 = 1$ for $p^t = p_{1'}^t$. Therefore, in $G^{mod}$, using the JKHE method, the c-color of the leftmost target point $p_{1'}^t$ is corrected by

$$
\begin{aligned}
C^c(p_{1'}^t) &= C^c(p_{1'}^t) + w_1 * f^{c,KBI}(p_{1'}^t, \mathcal{N}'(p_{1'}^t)) \\
&\quad + w_2 * f^{c,HE}(p_{1'}^t) \\
&= C^c(p_{1'}^t) + 1 * f^{c,KBI}(p_{1'}^t, \mathcal{N}'(p_{1'}^t))
\end{aligned}
\tag{9}
$$

From Eq. 9, the c-color of the leftmost target point in $G^{mod}$ is corrected by the KBI method (see Eq. 2) which has been used to correct color for the rightmost target point in $G^{cl}$. It indicates that the grouping-effect free property at the boundary between $G^{cl}$ and $G^{mod}$ holds.

We further show the grouping-effect free property at the boundary between $G^{mod}$ and $G^{dist}$ in our algorithm. The proving key is to show whether the color of the rightmost target point in $G^{mod}$ is corrected by the HE method which is used by correcting color for the leftmost target point in $G^{dist}$. On the contrary, in the sorted target point sequence $\langle p_{1'}^t, p_{2'}^t, ..., p_{|G^{mod}|'}^t \rangle$, the average distance between the rightmost target point $p_{|mod|'}^t$ and its valid reference aligned source point in $N'(p^t)$ is maximal among the $|G^{mod}|$ average distances. From Eq. 7, it leads to $w_1 = 0$ and $w_2 = 1$ for $p^t = p_{|mod|'}^t$. Therefore, in $G^{mod}$, the c-color of the rightmost target point $p_{|mod|'}^t$ is corrected by

$$
C^c(p_{|G^{mod}|'}^t) = C^c(p_{|G^{mod}|'}^t) + 1 * f^{c,HE}(p_{|G^{mod}|'}^t)
\tag{10}
$$

From Eq. 10, we observe that the c-color of the rightmost target point in $G^{mod}$ is corrected by the HE method which is used to correct color for the leftmost target point in $G^{dist}$. Consequently, by Definition 1, the grouping-effect free property at the boundary of $G^{mod}$ and $G^{dist}$ holds.

# 4 Experimental results

To compare the quantitative and qualitative quality performance of our color correction algorithm with the state-of-the-art methods, such as the NN method, the KNN method, the HM method, the HHM method, and the AGL method, large scale testing point cloud pairs are used. Finally, the ablation study is provided.

## 4.1 Testing point cloud pairs used and parameter setting

### 4.1.1 Testing point cloud pairs used

To compare the quantitative and qualitative quality performance of the considered color correction methods, the SHOT dataset and the augmented ICL-NUIM dataset are used as the two testing datasets. In our experiments, these 1086 testing color point cloud pairs cover different scenarios with high and low overlapping rates. Among these testing pairs, 146 point cloud pairs are collected from the SHOT dataset and 940 point cloud pairs are collected from the augmented ICL-NUIM dataset.

Table 1 lists the detailed information of the 176 point cloud pairs collected from the SHOT dataset. Among these 176 pairs, 76 point cloud pairs are with high overlapping rates and 70 point cloud pairs are with low overlapping rates. To create the 76 high overlapping rate pairs, 11, 11, 15, 13, 13, and 13 pairs are created from the Mario, Squirrel, PeterRabbit, Doll, Duck, and Frog testing sets, respectively, by collecting two consecutive point clouds in each testing set. To create the 70 low overlapping rate point cloud pairs, 10, 10, 14, 12, 12, and 12 pairs are created from the above-mentioned six testing sets by collecting two consecutive point clouds, with one frame skipped in between.

Table 2 lists the information of the 940 testing point cloud pairs collected from the augmented ICL-NUIM dataset. Among these 940 pairs, 226 point cloud pairs are with high overlapping rates and 714 point cloud pairs are with low overlapping rates. For quality comparison fairness, we first calculate the mean qualities of the color-corrected target points in the point cloud pairs with high overlapping rates and low overlapping rates, respectively, and then we average the two calculated mean qualities as the final quality of each considered color correction method. To create the 226 high overlapping rate pairs, 55, 75, 45, and 51 pairs are created from the Living Room 1, Living Room 2, Office 1, and Office 2 testing sets, respectively. To create the 714 low overlapping rate point cloud pairs, 272, 124, 186, and 132 pairs are created from the above-mentioned four testing sets. Note that to distinguish the high overlapping rate pairs from the low overlapping rate pairs, for each testing pair, we first apply the ground truth registration provided by the dataset to calculate the overlapping rate of that testing pair, and then compare the calculated overlapping rate with the specified threshold for determining the overlapping rate type of the testing pair.

**Table 1** The testing point cloud pairs collected from the SHOT dataset.

|              | #(point cloud pairs with high overlapping rates) | #(point cloud pairs with low overlapping rates) |
| ------------ | :---: | :---: |
| Mario        | 11 | 10 |
| Squirrel     | 11 | 10 |
| PeterRabbit  | 15 | 14 |
| Doll         | 13 | 12 |
| Duck         | 13 | 12 |
| Frog         | 13 | 12 |
| Total pairs  | 76 | 70 |

**Table 2** The testing point cloud pairs collected from the augmented ICL-NUIM dataset.

|               | #(point cloud pairs with high overlapping rates) | #(point cloud pairs with low overlapping rates) |
| ------------- | :---: | :---: |
| Living Room 1 | 55  | 272 |
| Living Room 2 | 75  | 124 |
| Office 1      | 45  | 186 |
| Office 2      | 51  | 132 |
| Total pairs   | 226 | 714 |

### 4.1.2 Parameter setting

For comparison fairness, the parameter setting in the comparative methods is configured according to the specifications set in their methods. In the proposed group partition strategy used in the proposed algorithm, the specified overlapping rate threshold $T_r$ is set to 45% and the specified distance threshold $T_d$ is set to 0.003 (meter).

## 4.2 Quantitative and qualitative quality comparison

To compare the color correction effects of the considered methods, the quantitative and qualitative quality performances are evaluated. Because there are no ground truth color corrected target point clouds provided in the testing datasets, the aligned source point clouds are taken as the ground truth point clouds. The popular quantitative quality metrics, such as color mean difference (CMD) and color peak signal-to-noise ratio (CPSNR), are used to evaluate the quantitative quality of each considered method.

After presenting the quantitative and qualitative quality comparison for the considered methods, the ablation study is provided.

### 4.2.1 Quantitative quality comparison and discussion

#### 4.2.1.1 Quantitative quality comparison

The two used quality metrics, namely CMD and CPSNR, are defined below. Given one aligned source point cloud $P^{s,ali}$ and the corresponding target point cloud $P^t$, after performing one considered color correction method on $P^{s,ali}$ and $P^t$, let $P^{t,cor}$ be the color-corrected target point cloud. Based on a variant of the color difference metric [34], the CMD metric of $P^{t,cor}$ is measured as

$$CMD = |\mu(P^{t,cor}) - \mu(P^{s,ali})|$$
$$= \frac{1}{3} \sum_{c \in \{R,G,B\}} |\mu^c(P^{t,cor}) - \mu^c(P^{s,ali})| \quad (11)$$

where for $c \in \{R, G, B\}$, $\mu^c(P^{t,cor})$ and $\mu^c(P^{s,ali})$ denote the c-color mean values of $P^{t,cor}$ and $P^{s,ali}$, respectively. The smaller the CMD value of one considered color correction method is, the better the color mean preservation of that method is.

The peak signal-to-noise ratio (PSNR) metric is originally used in image quality assessment. As the second quality metric, to evaluate the quality of the color corrected target point cloud $P^{t,cor}$, CPSNR is defined by

$$CPSNR = \frac{1}{|P^{t,cor}|} \sum_{i=1}^{|P^{t,cor}|} 10 \log_{10} \frac{255^2}{CMSE} \quad (12)$$

where the color mean square error (CMSE) between the target point $p_i^t (\in P^{t,cor})$ and the nearest aligned source point $p_i^{s,ali} (\in P^{s,ali})$ is expressed as $CMSE = \frac{1}{3} \sum_{c \in \{R,G,B\}} (C^c(p_i^{t,cor}) - C^c(p_i^{s,ali}))^2$.

In the first experiment design, there are two testing sets collected from the SHOT dataset, namely the set $T^h$ with 76 testing point cloud pairs with high overlapping rates and the set $T^l$ with 70 testing point cloud pairs with low overlapping rates. The average CMD values, denoted by $CDM^h$ and $CMD^l$, for the testing set $T^h$ and the testing set $T^l$ are obtained by averaging the 76 CMD and CPSNR values of $T^h$ and the 70 CMD and CPSNR values of $T^l$, respectively.

Table 3 tabulates the quantitative quality comparison among the considered methods. The fourth column of Table 3 indicates that the average CMD ($CMD^{avg}$) values of the NN method, the KNN method, the HM method, the AGL method, the HHM method, and our algorithm called "Ours" are 6.730, 6.732, 6.644, 7.143, 6.613, and 6.541, respectively. These $CMD^{avg}$ values indicate that our algorithm has the lowest average CMD value marked in black bold, and the HHM method [8] has the second-lowest average CMD value marked in orange bold.

As for the CPSNR quality comparison, the seventh column of Table 3 indicates that the average CPSNR ($CPSNR^{avg}$) values of the KNN, HM, AGL, HHM, and "Ours" methods are 41.598, 22.047, 14.527, 22.050, and 37.621, respectively, indicating that our algorithm achieving the second-best CPSNR performance.

**Table 3** The CMD and CPSNR comparison for the SHOT dataset.

| method | $CMD^h$ | $CMD^l$ | $CMD^{avg}$ | $CPSNR^h$ | $CPSNR^l$ | $CPSNR^{avg}$ |
|---|---|---|---|---|---|---|
| NN | 4.859 | 8.600 | 6.730 | - | - | - |
| KNN | 4.852 | 8.613 | 6.732 | 41.654 | 41.536 | **41.598** |
| HM | 5.013 | 8.274 | 6.644 | 22.988 | 21.024 | 22.047 |
| AGL | 5.524 | 8.761 | 7.143 | 14.688 | 14.352 | 14.527 |
| HHM | 4.930 | 8.295 | 6.613 | 22.981 | 21.039 | 22.050 |
| Ours | 4.723 | 8.359 | **6.541** | 32.317 | 43.379 | 37.621 |

In the second experiment design, as mentioned in Table 2, there are two testing sets, namely the set $T^h$ with high overlapping rates and the set $T^l$ with low overlapping rates. The average quality values for the testing set $T^h$ and the testing set $T^l$ are obtained by averaging the 226 quality values of $T^h$ and the 714 CMD values of $T^l$, respectively.

Based on the augmented ICL-NUIM dataset, Table 4 tabulates the quantitative comparison among the considered methods. The fourth column of Table 4 shows that the average CMD values of the NN method, the KNN method, the HM method, the AGL method, the HHM method, and our algorithm "Ours" are 9.992, 9.826, 9.840, 10.533, 9.747, and 9.648, respectively, indicating that our algorithm has the best average quantitative quality performance and the HHM method achieves the second-lowest average quantitative quality performance.

The seventh column of Table 4 indicates that the average CPSNR values of the KNN, HM, AGL, HHM, and "Ours" methods are 32.894, 21.910, 15.281, 21.917, and 28.896, respectively, also indicating that our algorithm achieving the second-best CPSNR performance.

**Table 4** The CMD and CPSNR comparison for the augmented ICL-NUIM dataset.

| method | $CMD^h$ | $CMD^l$ | $CMD^{avg}$ | $CPSNR^h$ | $CPSNR^l$ | $CPSNR^{avg}$ |
|---|---|---|---|---|---|---|
| NN | 8.241 | 11.744 | 9.992 | - | - | - |
| KNN | 8.179 | 11.473 | 9.826 | 32.906 | 32.891 | **32.894** |
| HM | 8.341 | 11.338 | 9.840 | 23.000 | 21.564 | 21.910 |
| AGL | 8.590 | 12.476 | 10.533 | 15.637 | 15.168 | 15.281 |
| HHM | 8.263 | 11.230 | 9.747 | 22.989 | 21.577 | 21.917 |
| Ours | 8.083 | 11.214 | **9.648** | 26.028 | 29.804 | 28.896 |

Based on the $CMD^{avg}$ and $CPSNR^{avg}$ values in Tables 3-4, the average quality comparison results are tabulated in Table 5. Table 5 shows that our algorithm has the best $CMD^{avg}$ performance and the second-best $CPSNR^{avg}$ performance; the HHM method has the second-best $CMD^{avg}$ performance and the KNN method has the best $CPSNR^{avg}$ performance. In average, our algorithm achieves the best quantitative quality performance.

**Table 5**  The average quality comparison for the two used datasets.

| method | $CMD^h$ | $CMD^l$ | $CMD^{avg}$ | $CPSNR^h$ | $CPSNR^l$ | $CPSNR^{avg}$ |
|--------|---------|---------|-------------|-----------|-----------|---------------|
| NN     | 6.550   | 10.172  | 8.361       | -         | -         | -             |
| KNN    | 6.515   | 10.043  | 8.279       | 35.107    | 33.662    | **34.064**    |
| HM     | 6.677   | 9.806   | 8.242       | 22.997    | 21.516    | 21.928        |
| AGL    | 7.057   | 10.619  | 8.838       | 15.398    | 15.095    | 15.179        |
| HHM    | 6.597   | 9.763   | *8.180*     | 22.987    | 21.529    | 21.935        |
| Ours   | 6.403   | 9.787   | **8.095**   | 27.610    | 31.016    | *30.069*      |

### 4.2.2 Qualitative quality comparison

The qualitative quality merit of our algorithm is presented first, and then the grouping-effect free property of our algorithm is demonstrated via the boundary-region comparison.

In the first experiment design, three testing point cloud pairs, one pair with a high overlapping rate, another pair still with a high overlapping rate, and the third pair with a low overlapping rate, are used to justify the visual effects of the color-corrected target point clouds of our algorithm. The perceptual merits of our algorithm for the other 143 point cloud pairs can be accessed from the website: `https://github.com/ivpml84079/Point-cloud-color-correction`.

Fig. 5 illustrates the color correction results using the considered methods. Fig. 5(a) illustrates the first testing point cloud pair "Mario" with a high overlapping rate, where the source point cloud is shown at the left side and the target point cloud is shown at the right side. Fig. 5(b) illustrates the alignment of Fig. 5(a). The right white glove in Fig. 5(b) is amplified and shown at the lower side of Fig. 5(c); the left white glove and the left ear in Fig. 5(b) are amplified and shown at the upper side of Fig. 5(c).

When comparing the perceptual results in Figs. 5(d)-(i), we observe that for the occluded left white glove and left ear within the blue rectangle, the qualitative quality performance of the HM method, the AGL method, the HHM method, and our algorithm are clearly superior to the NN method and the KNN method because of the lack of consideration for the global reference aligned source points in the NN and KNN methods. For the right white glove within the green rectangle, the qualitative quality performance of the HM method, the AGL method, and the HHM method are clearly inferior to the NN and KNN methods and our algorithm because of the lack of consideration for the local reference aligned source points in the HM, AGL, and HHM methods. As demonstrated in the lower side of Fig. 5(i), the grey color on the noisy portion of the right glove has been well corrected using our algorithm; as demonstrated in the upper side of Fig. 5(i), both of the color of the occluded left glove and the color of the left ear have also been well corrected using our algorithm.

Fig. 6 illustrates the visual merit of our algorithm for the second point cloud pair "duck" with a middle overlapping rate in Fig. 6(a). Fig. 6(b) illustrates the alignment result. The duck's beak and wing in Fig. 6(b) are amplified and illustrated at the top and bottom sides of Fig. 6(c), respectively. From the results in Figs. 6(d)-(i), we observe that for the duck's wing within the green rectangles, the qualitative quality effects of the HM method, the HHM method, and our algorithm are superior to the NN, AGL, and KNN methods. For the duck's beak within the blue rectangles, the HM and HHM methods produce visually unpleasant spots, but our algorithm achieves good visual results. Overall, our algorithm has the best visual performance relative to the comparative methods.

Fig. 7 illustrates the visual merit of our algorithm for the third point cloud pair "Frog" with a low overlapping rate in Fig. 7(a). Fig. 7(b) illustrates the alignment result. The potted plant in Fig. 7(b) is amplified and shown at the left side of Fig. 7(c); the frog's back and the potted plant in Fig. 7(b) are amplified and shown at the right side of Fig. 7(c). After comparing the perceptual results in Figs. 7(d)-(i), we observe that for the frog's back and the potted plant within the green rectangle, the qualitative quality effects of the HM, AGL, and HHM methods, and our algorithm are superior to the NN and KNN methods because of the lack of consideration for the global reference aligned source points in NN and KNN. For the potted plant within the blue rectangle, the qualitative quality effects of the HM, AGL, and HHM methods are also inferior to the NN method, the KNN method, and our algorithm because of the lack of consideration for the local reference aligned source points in HM, AGL, and HHM. As demonstrated in the left side of Fig. 7(i), the color on the noisy portion of the potted plant have been well corrected using our algorithm; as demonstrated in the right side of Fig. 7(i), the color of the occluded frog's back and the potted plant have also been well corrected using our algorithm. Overall, our algorithm has the best visual performance.

In the second experiment design, in the 940 testing point cloud pairs of the augmented ICL-NUIM dataset, the three selected pairs contain one pair with high overlapping rate, another pair with middle overlapping rate, and the other pair with a low overlapping rate. The perceptual merits of our algorithm for the remaining 937 point cloud pairs can also be checked on the aforementioned website: `https://github.com/ivpml84079/Point-cloud-color-correction`.

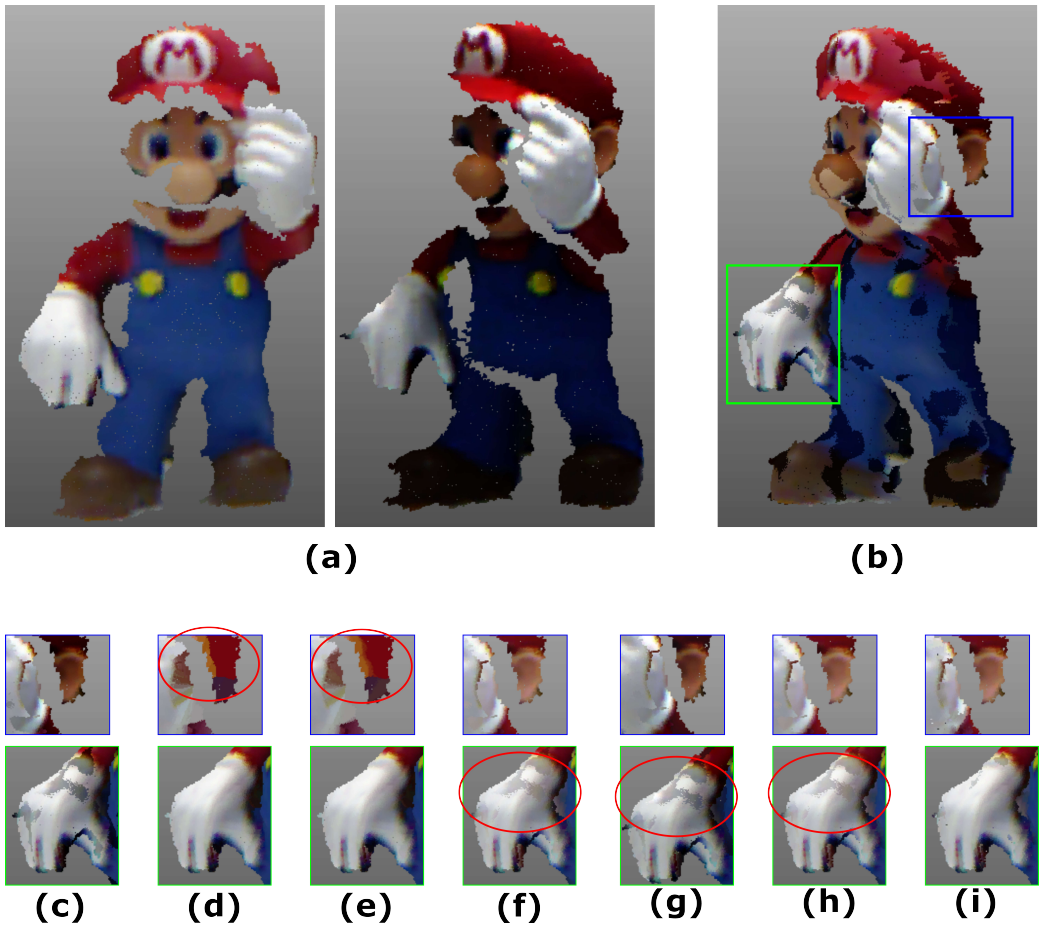

**(a)**    **(b)**

**(c)    (d)    (e)    (f)    (g)    (h)    (i)**

**Fig. 5** The perceptual quality merit of our algorithm for the first point cloud pair "Mario" with a high overlapping rate. (a) The input point cloud pair with source on the left and target on the right. (b) The alignment. (c) Two amplified regions cut off from Fig. 5(a). (d) NN. (e) KNN. (f) HM. (g) AGL. (h) HHM. (i) Ours.

Fig. 8(a) illustrates the first testing point cloud pair "Living Room 1" with a high overlapping rate. Fig. 8(b) illustrates the alignment of Fig. 8(a). The brick wall in Fig. 8(b) is amplified and shown at the lower side of Fig. 8(c); the shadow on the wall in Fig. 8(b) is amplified and shown at the upper side of Fig. 8(c).

After checking the perceptual results in Figs. 8(d)-(i), we observe that for the occluded brick wall within the green rectangle, the qualitative quality effects of the HM method, the AGL method, the HHM method, and our algorithm are clearly superior to the NN method and the KNN method because of the lack of consideration for the global reference aligned source points in the NN and KNN methods. For the shadow on the wall within the blue rectangle, the qualitative quality effects of the HM method, the AGL method, and the HHM method are inferior to the NN and KNN methods and our algorithm because of the lack of consideration for the local reference aligned source points in the HM, AGL, and HHM methods. As demonstrated in the lower side of Fig. 8(i), the occluded texture of the brick wall has been well corrected using our algorithm; as demonstrated in the upper side of Fig. 8(i), the color of the shadow on the wall has also been well corrected using our algorithm. Overall, our algorithm has the best visual performance for Fig. 8(b).

Fig. 9 illustrates the visual merit of our algorithm for the second point cloud pair "Living Room 2" with a middle overlapping rate in Fig. 9(a). Fig. 9(b) illustrates the alignment result. The sofa and the desk in Fig. 9(b) are amplified and illustrated at the top and bottom sides of Fig. 9(c), respectively. From the results in Figs. 9(d)-(i), we observe that for the sofa within the blue rectangles, the qualitative quality effects of the HM method, the HHM method, and our algorithm are superior to the NN, AGL, and KNN methods. For the desk within the green rectangles, the HM and HHM methods produce visually unpleasant spots near the table leg, but our algorithm achieves good visual results. Overall, our algorithm has the best visual performance relative to the comparative methods.

Fig. 10 illustrates the color correction results for the third point cloud pair "Office 1" with a low overlapping rate. Fig. 10(b) illustrates the alignment result. The patterned floor in Fig. 10(b) is amplified and shown at the lower side of Fig. 10(c); the corner of the wall in Fig. 10(b) is amplified and shown at the upper side of Fig. 10(c). After comparing the perceptual results in Figs. 10(d)-(i), we observe that for the occluded patterned floor within the green rectangle, the visual quality effects of the HM, AGL, and HHM methods, and our algorithm are superior to the NN and KNN methods because of the lack of consideration for the global reference aligned source points in NN and KNN. For the corner of the wall within the blue rectangle, the visual quality effects of the HM, AGL, and HHM methods are inferior to the NN and KNN methods and our algorithm because of the lack of consideration for the local reference aligned source points in the HM, AGL, and HHM methods. As demonstrated in the lower side of Fig. 10(i), the color on the noisy portion of the occluded patterned floor has been well corrected using our algorithm; as demonstrated in the upper side of Fig. 10(i), the color of the corner of the wall has also been well corrected using our algorithm. Overall, our algorithm has the best visual performance.

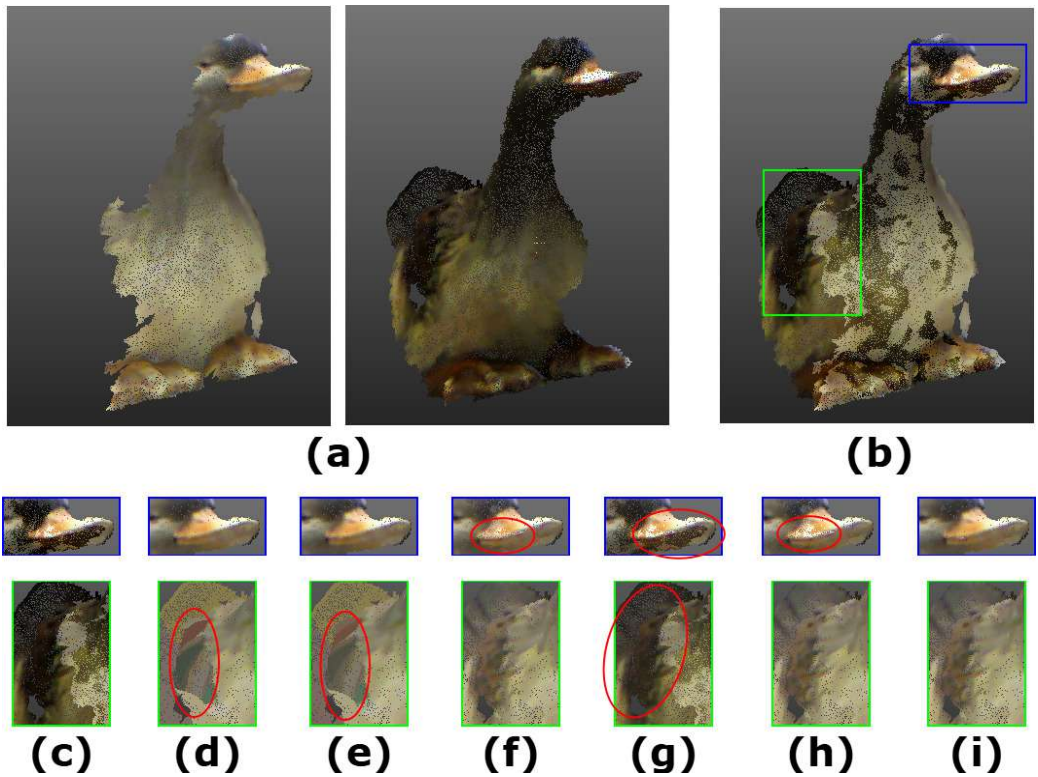

**Fig. 6** The perceptual quality merit of our algorithm for the second point cloud pair "duck" with a middle overlapping rate. (a) The input point cloud pair with source on the left and target on the right. (b) The alignment. (c) Two amplified regions cut off from Fig. 6(a). (d) NN. (e) KNN. (f) HM. (g) AGL. (h) HHM. (i) Ours.

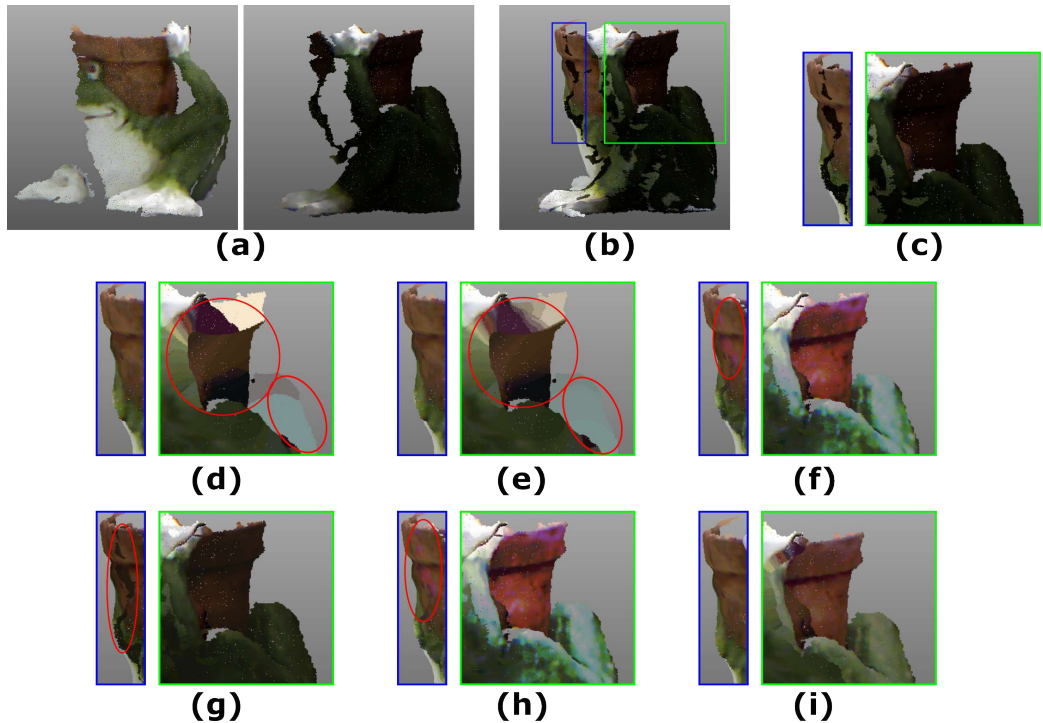

**Fig. 7** The perceptual quality merit of our algorithm for the third point cloud pair "Frog" with a low overlapping rate. (a) The input point cloud pair with source on the left and target on the right. (b) The alignment. (c) Two amplified regions cut off from Fig. 7(a). (d) NN. (e) KNN. (f) HM. (g) AGL. (h) HHM. (i) Ours.

Two amplified regions cut off from the duck head in Fig. 6(i) and the sofa chair surface Fig. 9(i) are illustrated to show the perceptual grouping-effect free advantage of the proposed algorithm. The left-hand image in Fig. 11(a) shows the three partitioned groups for the duck head in Fig. 6(i), where the pink, red, and dark red areas denote the point sets in $G^{cl}$, $G^{mod}$, and $G^{dist}$, respectively. The right-hand image in Fig. 11(a) shows the alignment result, and it is observed that the colors between two adjacent boundary groups are rather smooth. Similarly, The left-hand and right-hand images in Fig. 11(b) shows the three partitioned groups for the sofa chair surface and the alignment result in Fig. 9(i), respectively. From the right-hand image in Fig. 11(b), it is observed that the colors between two adjacent boundary groups are also smooth, highlighting the perceptual grouping-effect free advantage of the proposed algorithm.

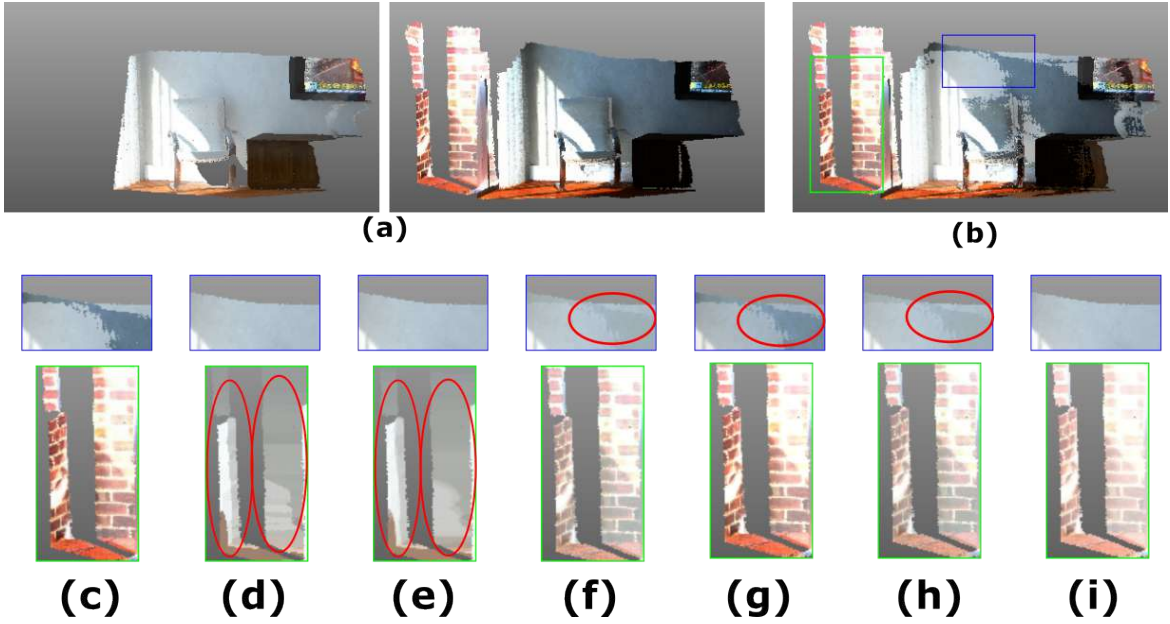

**Fig. 8** The perceptual quality merit of our algorithm for the second point cloud pair "Living Room 1" with a high overlapping rate. (a) The input point cloud pair with source on the left and target on the right. (b) The alignment. (c) Two amplified regions cut off from Fig. 8(a). (d) NN. (e) KNN. (f) HM. (g) AGL. (h) HHM. (i) Ours.

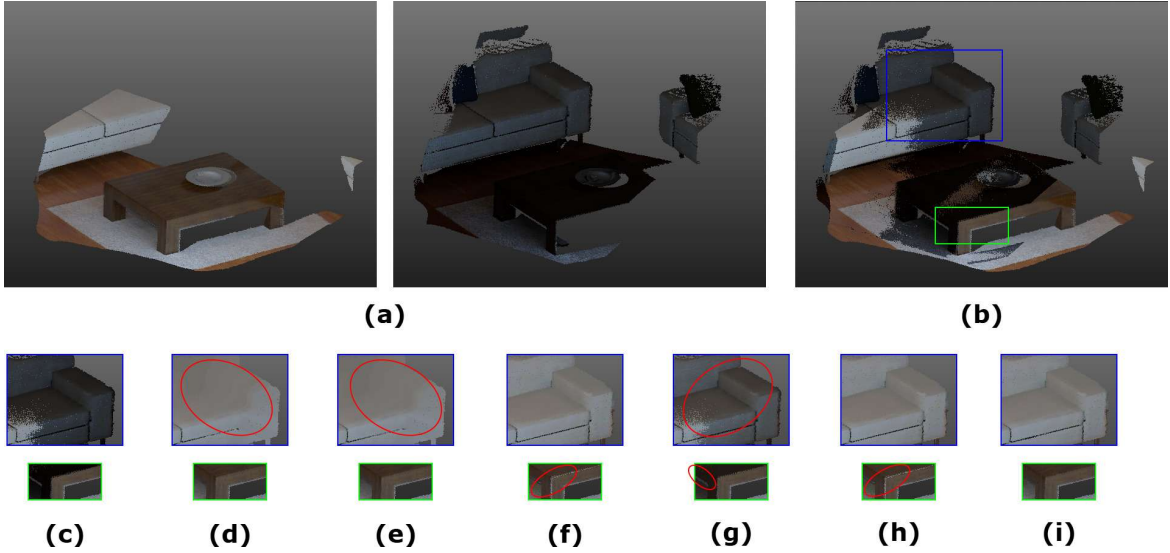

**Fig. 9** The perceptual quality merit of our algorithm for the second point cloud pair "Living Room 2" with a middle overlapping rate. (a) The input point cloud pair with source on the left and target on the right. (b) The alignment. (c) Two amplified regions cut off from Fig. 9(a). (d) NN. (e) KNN. (f) HM. (g) AGL. (h) HHM. (i) Ours.

### 4.2.3 Ablation study

The ablation study of our algorithm is set up by considering the importance of the bi-group/tri-group partition strategy focusing on KBI vs. JKHE; the bi-group/tri-group partition strategy focusing on weight tuning in JKHE.

Table 6 tabulates the ablation results for the bi-group/tri-group partition strategy focusing on KBI vs. JKHE. In the first combination, the bi-group case only considers the KBI method for $G^{cl}$ and $G^{mod}$; the tri-group case considers the KBI method for $G^{cl}$ and $G^{mod}$, but the HE method for $G^{dist}$. In the second combination, the bi-group case only considers the JKHE method for $G^{cl}$ and $G^{mod}$; the tri-group case considers the JKHE method for $G^{cl}$ and $G^{mod}$, but the HE method for $G^{dist}$. The third combination is exactly our color correction algorithm. From Table 6, relative to the first and second combinations, it is observed that the CMD and CPSNR gains are positive when using the third combination, i.e., our algorithm. In detail, relative to the first combination, our algorithm can improve the CMD value from 12.201 to 8.095 and can improve the CPSNR value from 28.488 to 29.413. Relative to the second combination, our algorithm can improve the CMD from 11.997 to 8.095 and can improve the CPSNR value from 23.074 to 29.413.

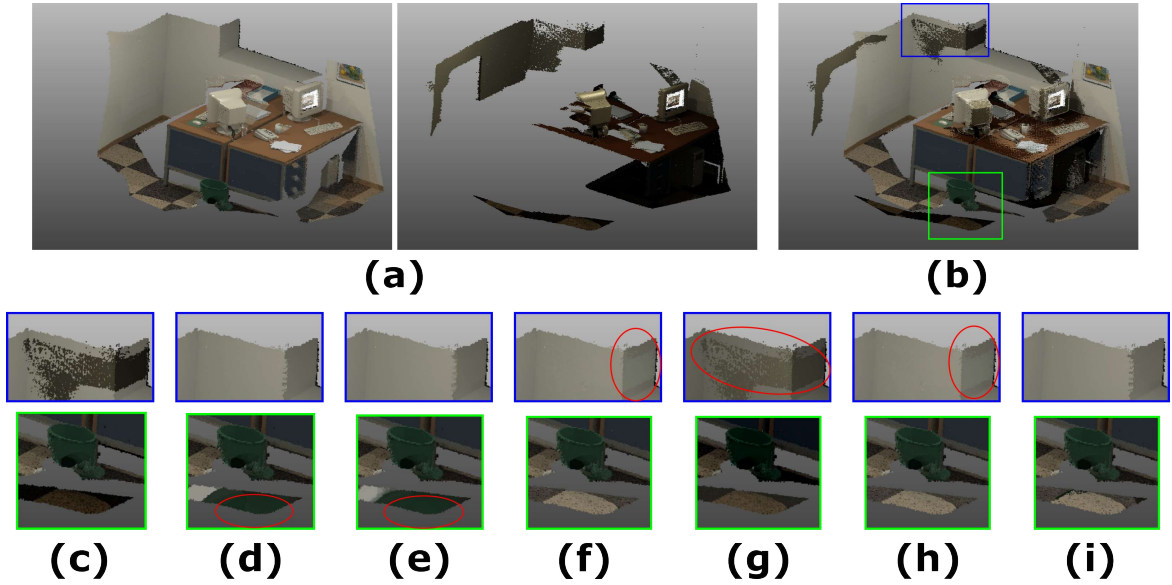

**Fig. 10** The perceptual quality merit of our algorithm for the third point cloud pair "Office 1" with a low overlapping rate. (a) The input point cloud pair with source on the left and target on the right. (b) The alignment. (c) Two amplified regions cut off from Fig. 10(a). (d) NN. (e) KNN. (f) HM. (g) AGL. (h) HHM. (i) Ours.

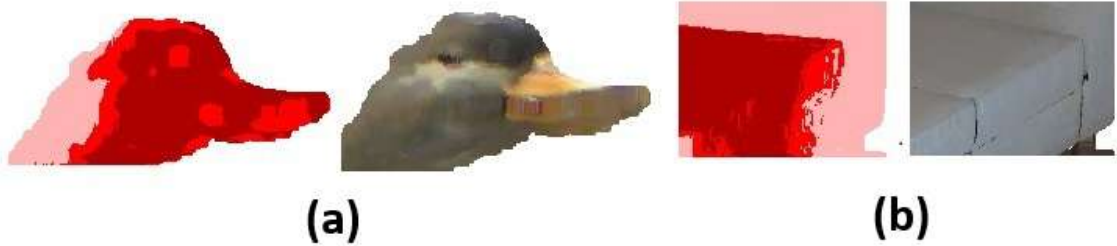

**Fig. 11** The perceptual grouping-effect free advantage of the proposed algorithm for the duck head in Fig. 6(i) and the sofa chair surface in Fig. 9(i). (a) The three partitioned groups and the alignment result for duck head. (b) The three partitioned groups and the alignment result for sofa chair surface.

Table 7 tabulates the ablation results for the bi-group partition strategy vs. the tri-group partition strategy and the weight tuning in JKHE. Five combinations are included. In the first combination, the target points in $P^t$ are only partitioned into two groups and the weight tuning method is used in JKHE. In the second combination, all target points in $P^t$ are only partitioned into two groups, but the two weights are assigned evenly, i.e., $w_1 = w_2 = \frac{1}{2}$, in JKHE. Relative to the first and second combinations, our algorithm, i.e., the fifth combination, can significantly improve the CMD value from 12.429 and 12.194 to 8.095, respectively, but our algorithm slightly degrades the CPSNR value from 29.573 and 29.567 to 29.413, respectively.

In the third combination, all target points are only partitioned into three groups and the weight tuning method is used in JKHE. In the fourth combination, all target points are also partitioned into three groups, but in JKHE, the two weights are assigned evenly, i.e., $w_1 = w_2 = \frac{1}{2}$. Relative to the third and fourth combinations, our algorithm can significantly improve the CMD value from 12.219 and 10.988 to 8.095, respectively, and our algorithm can improve the CPSNR value from 28.957 and 28.794 to 29.413, respectively.

**Table 6** Ablation study for KBI vs. JKHE.

| Tri-group | | | | |
| Bi-group | | | CMD | CPSNR |
| KBI | JKHE | HE | | |
|---|---|---|---|---|
| O | X | O | 12.201 | 28.488 |
| X | O | O | 11.997 | 23.074 |
| O | O | O | 8.095 | 29.413 |

**Table 7** Ablation study for two-group vs. three-group and weight tuning in JKHE.

| Bi-group | Tri-group | weight tuning in JKHE | CMD | CPSNR |
|---|---|---|---|---|
| O | X | O | 12.429 | 29.573 |
| O | X | X | 12.194 | 29.567 |
| X | O | O | 12.219 | 28.957 |
| X | O | X | 10.988 | 28.794 |
| O | O | O | 8.095 | 29.413 |

# 5 Conclusions

The proposed color correction algorithm has been presented for color point clouds. In our algorithm, the first novelty is the proposal of an overlapping rate-based partition method to determine a proper partition strategy, either the bi-group partition or the tri-group partition. The bi-group partition strategy partitions all target points into two groups: $G^{cl}$ and $G^{mod}$; the tri-group partition strategy partitions all target points into three groups: $G^{cl}$, $G^{mod}$, and $G^{dist}$. The second novelty of our algorithm is that for target points in $G^{cl}$, $G^{mod}$, and $G^{dist}$, the suitable color correction methods, KBI, JKHE, and HE, are proposed, respectively. The third novelty is the existence of the grouping-effect free property in our algorithm. We have conducted extensive experiments on 1086 testing color point cloud pairs to show the quantitative and qualitative quality merits of our algorithm when compared with the previous methods: the NN, KNN, HM, AGL, and HHM methods.

The first future work is to extend our color correction results from two point clouds to multiple point clouds, and the registration methods [29], [9], [33], [43] could be adopted to align these multiple point clouds. The second future work is to apply other machine learning methods, such as support vector machine, multilayer perceptron, decision tree, and autoencoders [10], [2], to design the grouping strategy used in our algorithm for improving the color correction performance.

**Acknowledgments** The authors appreciate the valuable comments provided by the two reviewers to improve the manuscript.

**Author Contributions** Kuo-Liang Chung: Conceptualization, Methodology, Experiments designed and conducted, Wrote the manuscript. Ting-Chung Tang: Software, Experiments conducted, Validation, Visualization, Analysis, Edited the manuscript.

**Funding** This work was supported by Grants MOST-110-2221-E-011-088-MY3 and Grant MOST-111-2221-E-011-126-MY3 of the Ministry of Science and Technology, Taiwan.

**Data availability** The data presented in this study are openly available in `https://github.com/ivpml84079/Point-cloud-color-correction`

## Declarations

**Conflict of interest** The authors declare that they have no conflicts of interest in this paper.

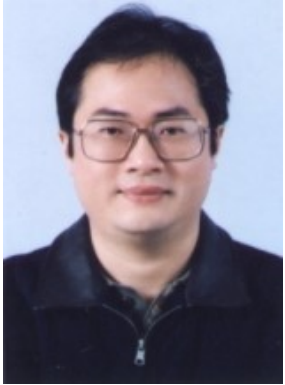

**Kuo-Liang Chung** Kuo-Liang Chung received his B.S., M.S., and Ph.D. degrees from National Taiwan University, Taipei, Taiwan, in 1982, 1984, and 1990, respectively. He has been a Chair Professor with the Department of Computer Science and Information Engineering at National Taiwan University of Science and Technology since 2009. He was a recipient of Distinguished Research Award (2004- 2007, 2019-2022), Distinguished Research Project Award (2009-2012), and Distinguished Research Fellow (2024) from the Ministry of Science and Technology, Taiwan. His research interests include image and point cloud processing. He has been an Associate Editor of the Journal of Visual Communication and Image Representation since 2011.

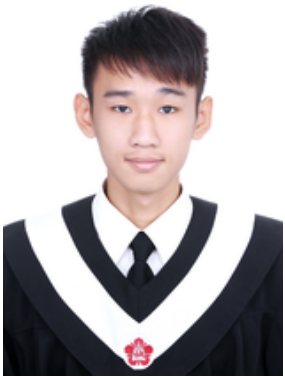

**Ting-Chung Tang** received his B.S. degree in Computer Science and Engineering from National Cheng Kung University, Tainan, Taiwan, in 2018. He received his M.S. degree in Computer Science and Information Engineering at National Taiwan University of Science and Technology, Taipei, Taiwan, in 2024. His research interests include point cloud processing, image processing and deep learning applications.